\documentclass{article}

\PassOptionsToPackage{numbers}{natbib}
\usepackage[final]{mlncp_2023}

\usepackage[utf8]{inputenc} 
\usepackage[T1]{fontenc}    
\usepackage{hyperref}       
\usepackage{breakurl}
\usepackage{url}            
\usepackage{booktabs}       
\usepackage{amsfonts}       
\usepackage{nicefrac}       
\usepackage{microtype}      
\usepackage{xcolor}         
\usepackage{csquotes}
\usepackage{todonotes}
\usepackage{siunitx}
\usepackage{amsmath}
\usepackage{amssymb}
\usepackage{amsthm}
\usepackage{multirow}
\usepackage{caption}
\usepackage{subcaption}
\usepackage{collcell}
\usepackage{verbatim}
\usepackage[subtle]{savetrees}
\usepackage{hyperref}
\hypersetup{
    colorlinks=true,
    linkcolor=black,
    filecolor=magenta,      
    urlcolor=blue,
    citecolor=black
}       

\usepackage{graphicx}

\newcommand{\bfa}{\mathbf{a}}
\newcommand{\bfu}{\mathbf{u}}
\newcommand{\bfr}{\mathbf{r}}
\newcommand{\bfs}{\mathbf{s}}

\newcommand{\bfc}{\mathbf{c}}
\newcommand{\bfx}{\mathbf{x}}
\newcommand{\bfy}{\mathbf{y}}
\newcommand{\bfz}{\mathbf{z}}

\newcommand{\bfth}{\boldsymbol{\vartheta}}

\newcommand{\bfW}{\ensuremath{\mathbf{W}}}

\newcommand{\bfb}{\ensuremath{\mathbf{b}}}

\newcommand{\dti}{^{\langle t \rangle}}
\newcommand{\dtim}[1]{^{\langle t-{#1} \rangle}}

\newcommand{\hp}{{\,\mathrel{\vcenter{\hbox{\tiny$\odot$}}}\,}}

\newcommand{\bwrap}[1]{\left( #1 \right)}
\newcommand{\concat}[1]{\left[ #1 \right]}
\newcommand{\fun}[2]{{#1}\bwrap{#2}}

\title{Activity Sparsity Complements Weight Sparsity for Efficient RNN Inference}

\author{%
  Rishav Mukherji$^1$\thanks{Work carried out at TUD Dresden University of Technology}
  \quad
  Mark Schöne$^2$
  \quad
  Khaleelulla Khan Nazeer$^2$
  \\
  \textbf{Christian Mayr}$^2$ 
  \quad
  \textbf{Anand Subramoney}$^3$
  \\
  $^1$ Birla Institute of Technology and Science, Pilani -- Goa Campus, India \\
  $^2$ TUD Dresden University of Technology, Germany 
  \quad
  $^3$ Royal Holloway, University of London, UK \\
  \texttt{rishavm16@gmail.com} \quad \texttt{mark.schoene@tu-dresden.de}
}

\newcolumntype{d}[1]{D{.}{.}{#1}}

\begin{document}

\maketitle

\begin{abstract}
Artificial neural networks open up unprecedented machine learning capabilities at the cost of ever growing computational requirements.
Sparsifying the parameters, often achieved through weight pruning, has been identified as a powerful technique to compress the number of model parameters and reduce the computational operations of neural networks.
Less well studied are sparse activations for computational efficiency, while omnipresent in both biological neural networks and deep learning systems.
Moreover, the interaction between sparse activations and weight pruning is not fully understood.
In this work, we demonstrate that activity sparsity can compose multiplicatively with parameter sparsity in a recurrent neural network model based on the GRU  that is designed to be activity sparse.
We achieve up to $20\times$ reduction of computation while maintaining  perplexities below \num{60} on the Penn Treebank language modeling task.
This magnitude of reduction has not been achieved previously with solely sparsely connected LSTMs, and the language modeling performance of our model has not been achieved previously with any sparsely activated recurrent neural networks or spiking neural networks.
Neuromorphic computing devices are especially good at taking advantage of the dynamic activity sparsity, and our results provide strong evidence that making deep learning models activity sparse and porting them to neuromorphic devices can be a viable strategy that does not compromise on task performance.
Our results also drive further convergence of methods from deep learning and neuromorphic computing for efficient machine learning.
\end{abstract}
\section{Introduction}
As the available compute per energy unit grows, artificial neural networks (ANNs) become increasingly popular for applications ranging from cloud services to mobile and edge systems.
While task performance is crucial for all applications, including on low-power environments such as mobiles, the energy consumption of the system is critical to allow deployment in such environments.
Many tasks have additional latency requirements for safety reasons or to enhance the user experience.
Hence, enhancing ANNs inference efficiency is vital for deep learning application deployment.

The power and latency of deployed neural networks depends on the number of memory accesses and the number of arithmetic and logic operations conducted by the system.
While on-chip SRAM access energy costs are comparable to arithmetic operations, DRAM access is orders of magnitude more energy and latency intensive \citep{Horowitz2014}.
Hence, the key performance indicators for accelerating neural network inference, energy and latency, are dominated by reading weights from DRAM.
This issue is intensified for the case of batch size 1 inference, a common setting for mobile applications, since the cost of fetching weights cannot be spread across multiple samples.

Compressing neural networks to reduce the energy and latency cost is an active area of research.
The most relevant techniques include (1.) sparse and low-rank weight matrices \citep{Hoefler2021}, and (2.) fixed-point integer quantization to commonly 8-bit and below \citep{Wu2020IntegerQF, Menghani2023}.
Sparse and low-rank matrices reduce the number of weights that have to be fetched from memory, while quantization reduces the number of bits transferred per weight.
Less popular is the topic of sparse neuron activations.
Sparse activations theoretically limit the weights that have to be fetched from memory to the columns/rows associated with the non-zero activations, a large potential efficiency gain.
Although, sparse activations have been observed in deep feed-forward networks \citep{kurtz20a, Hunter2022, li2023}, many hardware accelerators, with a few exceptions~\cite{Parashar2017, Yuan2020, Zhang2021}, do not leverage this type of sparsity due to their dynamic nature.
The ability to handle dynamic activation sparsity is a core feature of neuromorphic hardware such as \citep{Loihi2018, Mayr2019}.
Most neuromorphic systems operate in an event-driven manner, and co-locate memory and computation to reduce the energy and latencies~\citep{Christensen2022}.

In this work, we focus on activity sparse recurrent neural networks (RNNs) based on the recently published event-based GRU (EGRU) model~\citep{Subramoney2023}.
With the recent resurgence in RNNs with architectures that are able to get close to or beat transformers in language modeling task performance~\citep{orvieto2023resurrecting,dao2022hungry}, we expect our analysis to be relevant in a broader and more practical context as well.

Our contributions to this workshop are as follows:
\begin{itemize}
    \item Using an efficient recurrent architecture (EGRU) designed for novel neural network accelerators, we show how activity sparsity can be tuned using weight decay
    \item We show that the number of connections of small-scale language models can be compressed with minimal loss in perplexity, which previously has only been shown on subpar baseline models for the Penn Treebank and WikiText-2 datasets \citep{Zhu2018}.
    \item We demonstrate that the reduction factors from activity sparsity and weight sparsity compose multiplicatively to yield a significant reduction of required memory accesses and arithmetic operations of up to $20\times$ without compromising on perplexity.
\end{itemize}

\section{Related Work}
\label{sec:related-work}

\textbf{Pruning RNNs.} 
An extensive review of weight pruning techniques can be found in \citep{Hoefler2021}.
Pruning has been applied to a range of recurrent architectures including Elman RNNs, LSTMs, and GRUs \cite{Narang2017, Han2017, Zhu2018, Bellec2018}.
On speech recognition benchmarks compression rates of up to \SI{90}{\percent} can be achieved with the LSTM model without loss in performance \citep{Han2017, Zhu2018, Bellec2018}.
\citet{Dai2020} reported an increase of sparsity by expanding the linear transformations of LSTM gates to be multi layer neural networks.
The best pruned LSTM model for language modeling on the Penn Treebank dataset in the literature is reported by \cite{Zhu2018}. 
They achieve their best results, a perplexity of \num{77.5} (where lower is better), at a weight sparsity of \SI{80}{\percent}. 
We generally find that pruning has not been applied to more recent LSTM based language models such as the AWD-LSTM \cite{Merity2018}, which achieves a perplexity of \num{57.3}.

\textbf{Activity sparsity for RNNs.}
The ReLU activation function sparsifies RNN activations at the cost of unstable training dynamics.
\citet{Talathi2016, Li2018} addressed this by training diagonal RNNs with ReLU activations.
They traded non-linear recurrent operations for feed-forward operations, and used more parameters than required with shallow but fully connected RNNs.
Delta Networks \citep{Neil2017} operate on differences between hidden states at consecutive time-steps.
This operator is shown to be equivalent to standard RNNs, while the deltas can be thresholded to yield a sparse approximation.
Spiking neural networks (SNNs) based on the biologically plausible leaky-integrate-and-fire neuron hold the promise of efficient inference on neuromorphic hardware but struggle to achieve state-of-the-art performance on machine learning benchmarks \citep{Ghosh2009, Tavanaei2019, Taherkhani2020, Yamazaki2022}.
Aiming to bridge the efficiency promise of SNNs with the task performance of ANNs,
\cite{Wozniak2020} and \citet{Subramoney2023} combined deep neural network architectures with discontinuous step functions and state resets to mimic the behaviour of SNNs, while achieving better task performance.
\citet{Zhu2023} followed a similar approach and applied SNNs in the context of larger-scale language models.

\textbf{Combining activity sparsity and weight sparsity.} 
\citet{Hunter2022} introduced a structured sparse algorithm for top-k winner-takes-it-all activation sparsity for feed-forward networks. 
\citet{Gao2022} show significant efficiency gains of joint activity and weight sparsity with an FPGA-based LSTM accelerator on a speech recognition benchmark.
While weight pruning is a popular research topic in the context of SNNs \citep{Chen2018,Rathi2019, Nguyen2021, Chen2021, Kim2022, yin2023},
most works focus on image classification.

\textbf{Language modeling with RNNs.}
Small scale language modeling datasets such as Penn Treebank \citep{Penn1993} or WikiText-2 \citep{merity2017pointer} drove progress of LSTM based language models before \citet{Vaswani2017} enabled large-scale language modeling.
The best raw LSTM language models on the Penn Treebank dataset are reported in \cite{Merity2018} and \cite{melis2018on}.
Recently, a set of RNNs based on linear recurrences demonstrated strong results on large-scale language modeling tasks \cite{Gu2022, Fu2023, Peng2023}.
In this work, we work in the small data regime, and comparison with the newer architectures will be done in future work.


\section{Efficient Recurrent Neural Networks for Neuromorphic Accelerators}
With our method, we aim for two goals:
First, match the performance metrics targeted by the task.
Second, respect the power and compute budget of the hardware system.
We exploit lessons from deep learning as well as neuromorphic computing to significantly reduce the communication between processing elements and memory.

\subsection{Event-based Gated Recurrent Unit}
Long Short-Term Memory (LSTM) networks \citep{Hochreiter1997} and Gated Recurrent Units (GRU) \citep{Cho2014} allow long range learning by gating input and hidden state variables for better gradient propagation.
The Event-based GRU (EGRU)~\citep{Subramoney2023} combines gating mechanisms with spiking mechanisms inspired by biological neuron models.
Therefore, it distinguishes between a local cell state $\bfc = (c_1, \dots, c_n)$ and a communicated cell state $\bfy = (y_1, \dots, y_n)$, where $n$ is the hidden dimension.
In each time-step, the communicated state $\bfy$ is a sparsification of the local state $\bfc$ via the pointwise heaviside thesholding function
\begin{align}
y_i\dti \;=\; c_i\dti \, \fun{H}{ c_i\dti - \vartheta_i } \,, \quad \fun{H}{x} = \begin{cases} 1, \quad x \geq 0 \\ 0, \quad x<0 \end{cases} \, ,
\end{align}
where $\bfth = (\vartheta_1, \dots, \vartheta_n)$ is a potentially trainable threshold parameter.
This sparse state $\bfy$ is then passed to an update gate $\bfu$ and a reset gate $\bfr$, similar to the GRU model
\begin{equation}
\begin{split}
     \bfu\dti = \fun{\sigma}{ \bfW_{u}  \concat{ \bfx\dti,\; \bfy\dtim1 } + \bfb_u }\,, & \quad
     \bfr\dti = \fun{\sigma}{\bfW_{r} \concat{ \bfx\dti,\; \bfy\dtim1 } + \bfb_r } \, .
\end{split}
\end{equation}
The gates compute a proposed state $\bfz$ and the updated local state $\bfc$ as
\begin{equation}
\begin{split}
     \bfz\dti = \fun{g}{ \bfW_z \concat{ \bfx\dti,\;  \bfr\dti \hp\; \bfy\dtim1 } + \bfb_z }\,, & \quad
     \bfc\dti = \bfu\dti \hp\; \bfz\dti  + (1-\bfu\dti) \hp\; \bfc\dtim1 - \bfs\dti\, .
\end{split}
\end{equation}
Note that this is almost the update of the GRU, but with an additional reset term $\bfs\dti = \bfth \fun{H}{\bfc\dti - \bfth}$.
This term is motivated by the reset term commonly used in SNNs to improve activity sparsity (see \cite{Eshraghian2023} for a review).
Another common strategy is to attach surrogate gradients to the non-differentiable Heaviside function 
\(
    \frac{\mathrm{d}H}{\mathrm{d}c} = \lambda ~ \max
    \left(1 - \lvert c \rvert / \epsilon\right)
\)
similar to \cite{bellec_Long_2018}  to allow differentiation of the event-based system.

\subsection{Sparsely Connected Networks}
\label{sec:pruning}
Event-based systems such as EGRU improve efficiency by reducing the activity on each neuron-to-neuron channel.
An orthogonal method to reduce the communication of a system is removing neuron-to-neuron channels entirely, i.e. pruning weights of the neural network (see sec. \ref{sec:related-work}).
The most popular heuristic for weight removal is weight magnitude pruning \citep{Han2015}.
In weight magnitude pruning, we choose a set of target weight tensors. 
We, then, systematically identify the weights with the smallest magnitudes from the target tensors and remove a specified percentage by setting it to zero.
Following the recommendations in \cite{Hoefler2021}, we investigated various pruning routines. 
A two-step approach, wherein we first train the RNN model to convergence followed by sparsification through iterative pruning, produces the best results for our goals of inference performance and sparsity.

The specific pruning methodology we implement is a global unstructured weight magnitude pruning technique. 
At each step, we carry out weight magnitude pruning on all the weight tensors that constitute the RNN model. 
The weights to be pruned are selected globally from all the tensors except for the embedding vectors. 
By selecting the weights globally, we enable the layers that play a larger role in the forward pass to retain a commensurate proportion of its weights.
Our rationale for pruning the RNN weights and not the embeddings is based on the perception that RNN weights are more representative across different tasks rather than just language modeling.
After each pruning iteration, we allow the model to fine-tune for a few epochs before advancing to the subsequent pruning step. 
This iterative procedure is repeated until a pre-defined target sparsity level is achieved.
For further details on the pruning experimental setup, refer Appendix ~\ref{sup:Experiments}.

\subsection{Efficiency of Sparse Activations and Sparse Connectivity}
The efficiency gains of sparse activations and sparse weights complement each other in a multiplicative way.
This yields significantly more efficient systems compared to ones that have each of these sparsities separately.
Consider the linear transformation $\bfW \bfa$.
Let $\lambda_\text{a} = \mathbb{E}\left[ \bfa \neq 0 \right]$ denote the fraction of active neurons in each time step.
We then call $\sigma_\text{a} = 1 - \lambda_\text{a}$ the activation sparsity.
Likewise, we call $\sigma_\text{w}$ the weight sparsity of $\bfW$, and denote the fraction of non-zero connections $\lambda_\text{w}$.
The transformation $\bfW\bfa$ requires memory access and arithmetics for the non-zero weights $\bfW_{ij}$ for each non-zero $\bfa_j$.
Hence, we need to load $\lambda_\text{a}$ columns of $\bfW$ that each have a fraction of $\lambda_\text{w}$ non-zeros. 
Effectively the fraction of remaining operations compared to a dense vector matrix multiplication is $\lambda_\text{a} \cdot \lambda_\text{w}$.

\section{Results}
Our main goal for this work is to show that activity sparsity and weight sparsity can be combined for optimal inference efficiency on novel accelerators.
According to previous work (see  sec.~\ref{sec:related-work}), RNNs for language modeling in the computationally feasible domain are more sensitive to the removal of weights compared to RNNs for speech recognition. 
We, therefore, choose language modeling as the more challenging task to support our main claim and conducted the evaluation on the Penn Treebank \citep{Penn1993} and the WikiText-2 \citep{merity2017pointer} word-level language modeling datasets.
The task performance metric for both datasets is perplexity (i.e. exponentiated cross-entropy, where lower is better).
Since computational efficiency depends on hardware properties, there is no universal metric to quantify efficiency.
We choose multiply accumulate (MAC) operations as our metric, which poses a finegrained measure of theoretically required operations on digital hardware.

\begin{figure}
    \centering
    \includegraphics[width=1.0\textwidth]{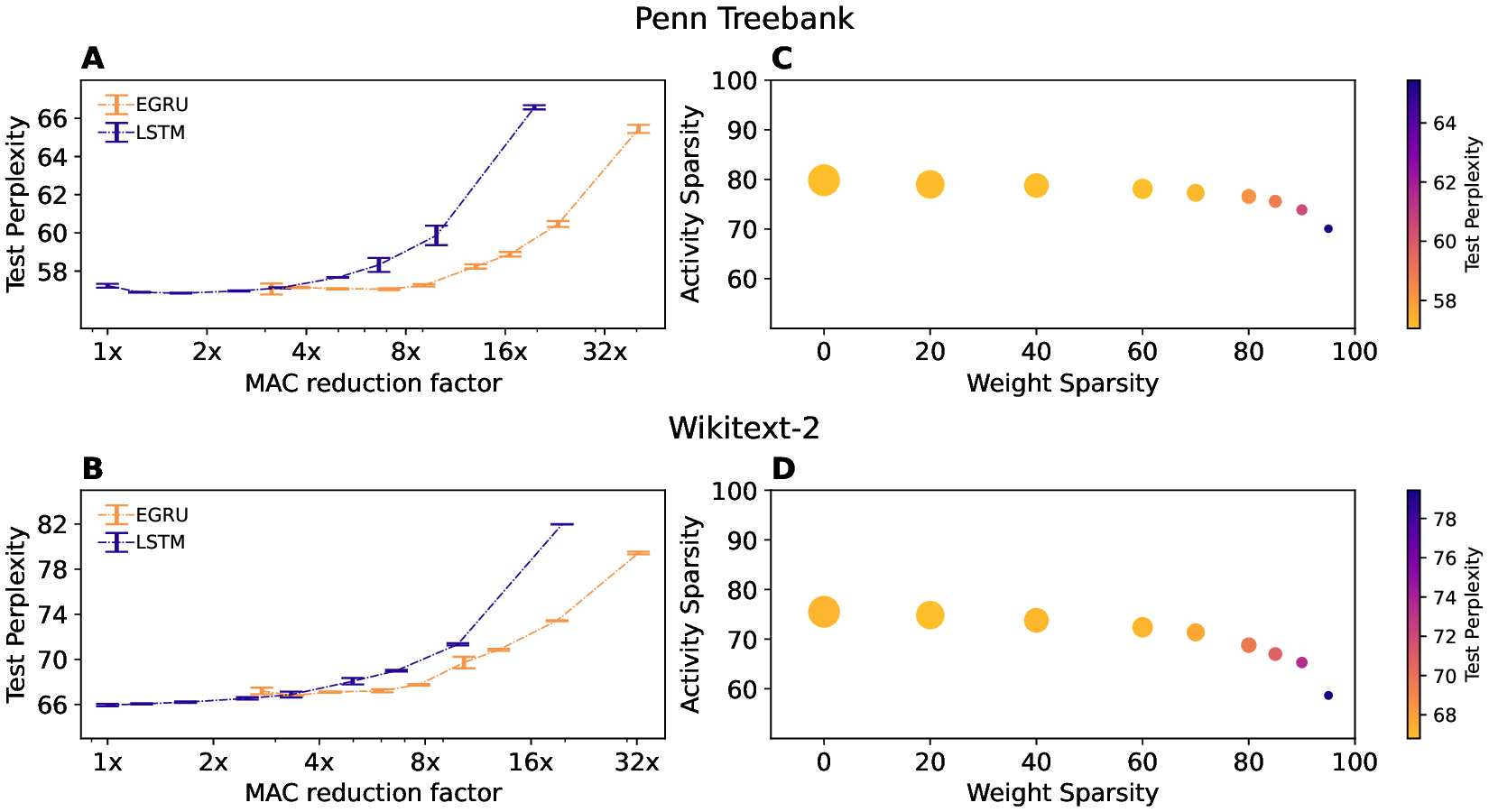}
    \caption{
    Influence of weight sparsity and activity sparsity on the Penn Treebank and WikiText-2 datasets.
    \textbf{A and B:} Test perplexity versus reduction of MAC operations through weight sparsity (LSTM) and combined activity and weight sparsity (EGRU). We plot the mean test perplexity and corresponding standard deviation over 15 seeds.
    \textbf{C and D:} Activity sparsity vs weight sparsity trade-off for EGRU. The marker size is proportional to the number of MAC operations while the colour represents task performance in terms of test perplexity.
    }
    \label{fig:Results}
\end{figure}

All models trained for this work follow the architecture presented in \cite{Merity2018}.
Hence, an embedding look-up table for the word embeddings is followed by three layers of stacked RNNs without skip connections and a linear decoder, whose weights are tied to the embedding layer.
DropConnect is applied to the recurrent weights \citep{Wan2013}.
In contrast to \cite{Merity2018}, we significantly simplify the optimization procedure by using AdamW instead of their proposed averaged SGD schedule.
AdamW speeds up the convergence of both \citet{Merity2018}'s LSTM and \citet{Subramoney2023}'s EGRU by a factor of 3-4.
While it would be natural to choose GRU as a baseline for comparison with EGRU, GRU models did not match the LSTM performance in our experiments.
This is consistent with the literature that does not report GRU results close to the LSTM baseline.
Since overfitting is a major problem on Penn Treebank and WikiText-2, we speculate that the sparse activations of EGRU may impose additional regularization of the model compared to GRU. 

Since this work focuses on the interaction of activity sparsity and weight sparsity, we do not use strategies such as mixture-of-softmaxes \citep{yang2018breaking}, neural cache \citep{Grave2017}, or mogrifier gates \citep{Melis2020Mogrifier} that can further improve the performance of the LSTM based methods.
We note that these methods could be applied on top of all the models presented in this work.

\subsection{Weight Sparsity}
We evaluate our weight pruning method across sparsity levels ranging from 20\% to 95\%. 
Training and subsequent pruning is carried out as per the methodology described in sec.~\ref{sec:pruning}. 
The results are presented in detail in tab.~\ref{table:Results}, 
and visualized in fig.~\ref{fig:Results}.
Weight magnitude pruning effectively compresses our models up to \SI{85}{\percent} with marginal loss in task performance on Penn Treebank.
 In fact, lightly pruned models, with weight sparsity levels up to \SI{60}\percent, actually exhibit an improvement in performance.
On WikiText-2, slightly lower compression rates are required to maintain task performance.
While a similar trend was observed by \citet{Zhu2018}, the results presented in tab.~\ref{table:Results} surpass their models significantly by \num{20} perplexity points. 

Keeping the MAC cost constant, if we compare a densely connected EGRU with sparse activity to a sparsely connected LSTM with dense activity, we find that they achieve comparable perplexities. However, combining both weight and activity sparsity yields better results than using either one in isolation.

%

\begin{table}
\centering
\begin{subtable}[c]{.5\linewidth}

\label{table:PTBResults}
\resizebox{!}{.28\linewidth}{
\begin{tabular}{l S[table-format=2.1] S[table-format=2.1] S[table-format=2.1] S[table-format=2.1]}
\toprule
\multirow{2}{2.5cm}{\\[-1.7ex] Model \\[0.2ex] (Weight sparsity)} & \multicolumn{2}{c}{LSTM} & \multicolumn{2}{c}{EGRU} \\
\cmidrule(lr){2-3} \cmidrule(lr){4-5}
& {MAC} & {Test PPL} & {MAC} & {Test PPL} \\
\midrule
\citet{Merity2018} & 20.2M & 57.3 & {-} & {-} \\
Ours ($ 0\%$)  &    20.2M &  57.1 & 6.4M & 56.6 \\
Ours ($20\%$)  &    16.2M &  56.9 & 5.3M & 57.1 \\
Ours ($40\%$)  &    12.1M &  56.8 & 4.1M & 56.9 \\
Ours ($60\%$)  &     8.1M &  56.9 & 2.8M & 57.0 \\
Ours ($70\%$)  &     6.1M &  57.1 & 2.2M & 57.1 \\
Ours ($80\%$)  &     4.1M &  57.6 & 1.6M & 58.0 \\
Ours ($85\%$)  &     3.1M &  57.7 & 1.2M & 58.7 \\
Ours ($90\%$)  &     2.0M &  58.3 & 0.9M & 60.2 \\
Ours ($95\%$)  &     1.0M &  66.5 & 0.5M & 65.2 \\
\bottomrule
\end{tabular}
}
\caption{Penn Treebank}
\end{subtable}%
\hfill
\begin{subtable}[l]{.5\linewidth}

\label{table:WT2Results}
\resizebox{!}{.28\linewidth}{
\begin{tabular}{l S[table-format=2.1] S[table-format=2.1] S[table-format=2.1] S[table-format=2.1]}
\toprule
\multirow{2}{2.5cm}{\\[-1.7ex] Model \\[0.2ex] (Weight sparsity)} & \multicolumn{2}{c}{LSTM} & \multicolumn{2}{c}{EGRU} \\
\cmidrule(lr){2-3} \cmidrule(lr){4-5}
& {MAC} & {Test PPL} & {MAC} & {Test PPL} \\
\midrule
\citet{Merity2018} & 20.2M & 65.8 & {-} & {-} \\
Ours ($ 0\%$)  &  20.2M &  65.7 &  7.4M &  66.6 \\
Ours ($20\%$)  &  16.2M &  65.9 &  6.0M &  66.7 \\
Ours ($40\%$)  &  12.1M &  66.1 &  4.7M &  67.0 \\
Ours ($60\%$)  &   8.1M &  66.4 &  3.3M &  67.1 \\
Ours ($70\%$)  &   6.1M &  66.3 &  2.6M &  67.6 \\
Ours ($80\%$)  &   4.1M &  68.0 &  2.0M &  69.4 \\
Ours ($85\%$)  &   3.1M &  68.9 &  1.6M &  70.7 \\
Ours ($90\%$)  &   2.0M &  71.2 &  1.1M &  73.3 \\
Ours ($95\%$)  &   1.0M &  81.9 &  0.6M &  79.3 \\
\bottomrule
\end{tabular}}
\caption{WikiText-2}
\end{subtable}%
\caption{Summarized results for the Penn Treebank and WikiText-2 datasets. We record the effective number of MAC operations in the RNN, expressed in millions. Lower value indicates greater efficiency. }
\label{table:Results}
\end{table}
\subsection{Activity Sparsity}
\label{sec:activity-sparsity}
\begin{figure}
    \centering
    \includegraphics[width=1.0\textwidth]{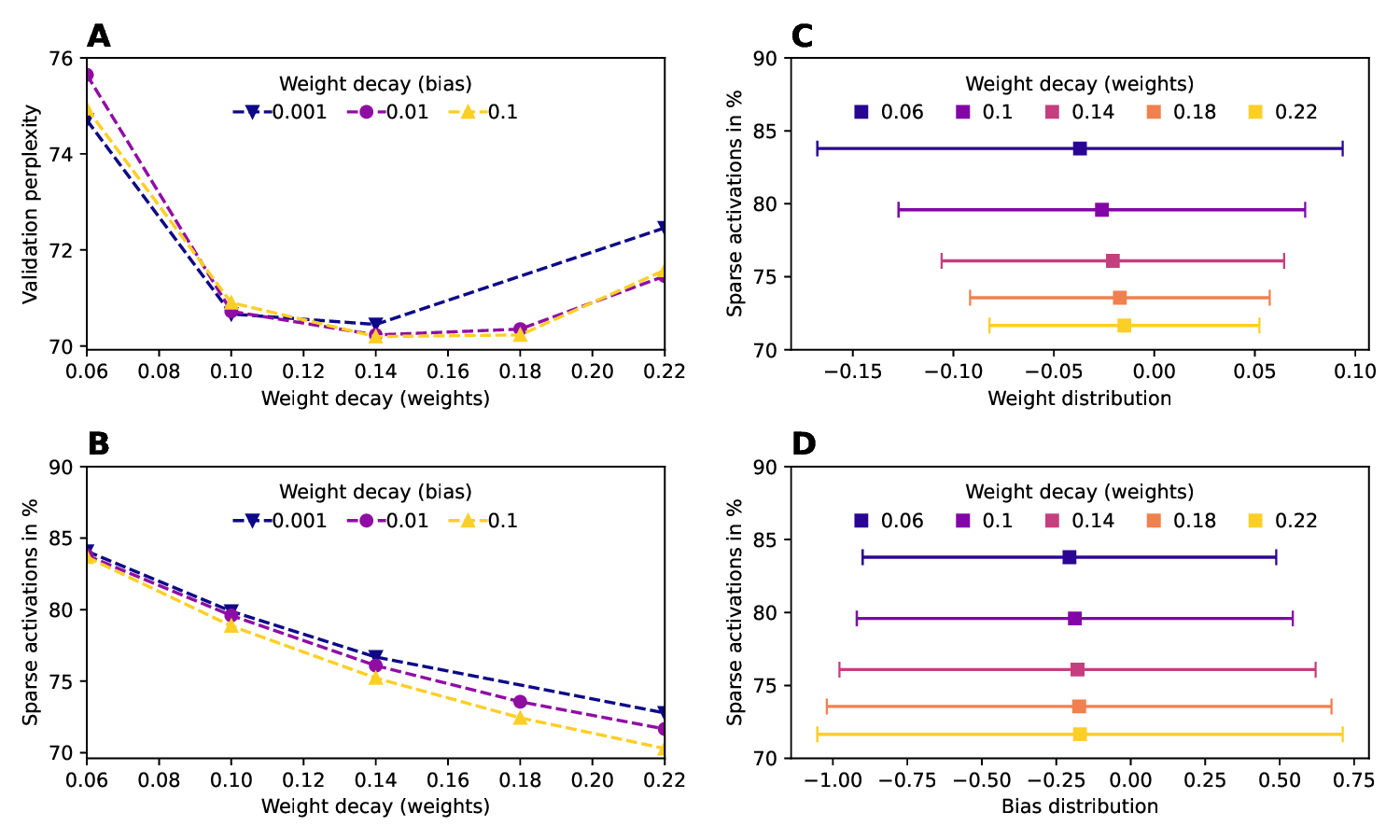}
    \caption{
    We show the effect of weight decay regularization on the performance and activity sparsity of the EGRU.
    Therefore, we consider separate degrees of weight decay for the weights and bias, separately.
    All models are trained on WikiText-2.
    \textbf{A:} Validation perplexity trade-off with weight decay.
    \textbf{B:} Weight decay on both the weights and the bias reduces the amount of sparse activations.
    \textbf{C} and \textbf{D}: Effect of weight decay on the distribution of weights and biases for fixed weight decay on the bias of \num{0.01}.}
    \label{fig:activity-sparsity}
\end{figure}
Utilizing \SI{85}{\percent} sparse weights and sparse activations, our method reduces the number of MAC operations by a theoretical factor of nearly \num{20} compared to the dense baseline set by \citet{Merity2018} on Penn Treebank.
This reduction in computational complexity is accompanied by a minimal perplexity gain of less than \num{2} points. 
Fig.~\ref{fig:Results}C and fig.~\ref{fig:Results}D show the trade-off between activity sparsity and weight sparsity for the EGRU model. 
It is discernible that there is minimal drop in activity sparsity as we increase weight sparsity until we reach high levels of weight sparsity; with the trade-off being slightly higher on WikiText-2. 
This observation provides further evidence that these two kinds of sparsities are mostly orthogonal and can be combined for greater efficiency in RNNs.

In our hyperparameter search, we observe that weight decay strongly influences both the task performance as well as the activity sparsity.
The effect of weight decay on the amount of sparse activations is particularly interesting as it provides a means to trade off sparsity for task performance.
We systematically study the influence of weight decay on the EGRU model by training a set of models with different degrees of weight decay applied to the weights and biases separately.
Fig. \ref{fig:activity-sparsity} summarizes our findings.
The task performance experiences an optimum around a weight decay of \num{0.14}.
Weight decay applied to the weights also has a strong influence on the activity of our model, while weight decay applied to the bias has a limited effect.
Investigating the distribution of weights and biases, we find that both weights and bias have a tendency to be negative, which drives the cell states below their threshold and promotes sparse activations. 
With stronger weight decay the distribution of weights concentrates closer around zero. 
Assuming statistical independence between weights $\bfW$ and activations $\bfa = \concat{ \bfx\dti,\; \bfy\dtim1 } $, the expectation of the preactivations is given by
$\mathbb{E}\left[ \bfW \bfa + \bfb \right] = \mathbb{E}\left[ \bfW \right] \mathbb{E}\left[ \bfa \right] + \mathbb{E}\left[ \bfb \right] \, .
$
Hence, negative mean weights tend to drive weights below thresholds, which increases the probability of preactivations passing the threshold $\bfth$ and reduces sparsity.

\section{Discussion}
This work shows that activity sparsity and connectivity sparsity complement each other for efficient recurrent neural network inference.
While the spiking neural network literature shows promising pruning results for image classification\citep{Rathi2019, Nguyen2021, Kim2022, yin2023}, SNNs do not yet deliver competitive baselines even for small-scale sequence modelling problems such as language modeling on Penn Treebank unlike more general event-based RNNs such as the EGRU~\citep{Subramoney2023}.
To the best of our knowledge this work is the first to show the multiplicative efficiency gain of activation sparsity and connectivity sparsity in the challenging domain of language modeling.

The EGRU model is just one example that shows the potential gains for both fields, i.e. improved benchmark performance for SNNs and improved efficiency for ANNs.
In the spirit of this workshop, our results suggest the need for more convergence of methods between deep learning and neuromorphic computing.
Such efforts will require joint commitment from accelerator designers and algorithm developers to explore models beyond the deep learning mainstream dominated by GPUs.
Unstructured weight sparsity alone does not necessarily justify the design of specific accelerators.
However, joint connectivity and activity sparsity could deliver the required reduction in operations that drive the efficiency of irregular accelerators beyond highly regular accelerators.

\section*{Acknowledgements}
 Rishav Mukherji was funded by the DAAD WISE scholarship for the duration of the work.
 Mark Schöne is fully funded by the Bosch Research Foundation.
 Khaleel Khan is funded by the German Federal Ministry of Education and Research (BMBF), funding reference 16ME0729K, joint project "EVENTS".
 The authors gratefully acknowledge the GWK support for funding this project by providing computing time through the Center for Information Services and HPC (ZIH) at TU Dresden.

\small

\bibliography{references}

\clearpage


\appendix
\normalsize
\section{Extended Results}
\label{sup:Results}

\begin{table}[!htb]
\caption{Penn Treebank results}
\label{table:PTB}
\begin{tabular}{c c c c c c c}
\toprule
\multirow{ 2}{*}{Model} & \multirow{ 2}{1cm}{\centering Weight \\[0.1ex] \centering Sparsity} & \multirow{ 2}{*}{MAC} & \multicolumn{2}{c}{Validation Perplexity} & \multicolumn{2}{c}{Test Perplexity}\\ \cmidrule(l){4-7}
  &  &  & Min &  Mean $\pm$ Std Dev &  Min &  Mean $\pm$ Std Dev \\ 
  \midrule
 \multirow{ 9}{*}{EGRU} &  $0\%$ & \tablenum{6.4 M} & \tablenum{60.81} &   $61.12 \pm 0.26$ & \tablenum{56.61} &  $57.06 \pm 0.29$ \\
  & $20\%$ &  \tablenum{5.3 M}& \tablenum{61.15} &  $61.22 \pm 0.04$ & \tablenum{57.10} &  $57.15 \pm 0.03$ \\
  & $40\%$ &  \tablenum{4.1 M }& \tablenum{61.01} &  $61.13 \pm 0.05$ & \tablenum{57.00} &  $57.07 \pm 0.03$ \\
  & $60\%$ &  \tablenum{2.8 M}& \tablenum{61.04} &  $61.12 \pm 0.04$ & \tablenum{57.01} &  $57.06 \pm 0.05$ \\
  & $70\%$ &  \tablenum{2.2 M}& \tablenum{61.22} &  $61.27 \pm 0.04$ & \tablenum{57.08} &  $57.25 \pm 0.06$ \\
  & $80\%$ &  \tablenum{1.6 M} & \tablenum{62.06} &  $62.31 \pm 0.15$ & \tablenum{58.04} &  $58.24 \pm 0.12$ \\
  & $85\%$ &  \tablenum{1.2 M} & \tablenum{62.76} &  $63.01 \pm 0.17$ & \tablenum{58.74} &  $58.88 \pm 0.12$ \\
  & $90\%$ &  \tablenum{0.9 M} & \tablenum{64.38} &  $64.66 \pm 0.22$ & \tablenum{60.22} &  $60.46 \pm 0.16$ \\
  & $95\%$ &  \tablenum{0.5 M} & \tablenum{69.64} &  $70.03 \pm 0.31$ & \tablenum{65.22} &  $65.44 \pm 0.21$ \\
 \midrule
 \multirow{ 9}{*}{LSTM} &  $0\%$ &   \tablenum[table-format = 2.1]{20.2M}& \tablenum{59.81} &  $60.04 \pm 0.18$ & \tablenum{57.06} &  $ 57.23 \pm 0.10 $ \\
  & $20\%$ &  \tablenum[table-format = 2.1]{16.2M }& \tablenum{59.61} &  $ 59.64 \pm 0.02$ & \tablenum{56.87} &  $ 56.89 \pm 0.02$ \\
  & $40\%$ &   \tablenum[table-format = 2.1]{12.1M}& \tablenum{59.53} &  $ 59.57 \pm 0.02$ & \tablenum{56.81} &  $ 56.84 \pm 0.02$ \\
  & $60\%$ &   \tablenum{ 8.1M}& \tablenum{59.58} &  $ 59.61 \pm 0.02$ & \tablenum{56.93} &  $ 56.96 \pm 0.03$ \\
  & $70\%$ &   \tablenum{ 6.1M}& \tablenum{59.73} &  $ 59.77 \pm 0.03$ & \tablenum{57.06} &  $ 57.12 \pm 0.03$ \\
  & $80\%$ &   \tablenum{ 4.1M}& \tablenum{60.28} &  $ 60.36 \pm 0.04$ & \tablenum{57.63} &  $ 57.67 \pm 0.02$ \\
  & $85\%$ &   \tablenum{ 3.1M}& \tablenum{61.04} &  $ 61.10 \pm 0.03$ & \tablenum{57.67} &  $ 58.32 \pm 0.37$ \\
  & $90\%$ &   \tablenum{ 2.0M}& \tablenum{62.88} &  $ 62.96 \pm 0.04$ & \tablenum{58.30} &  $ 59.87 \pm 0.51$ \\
  & $95\%$ &   \tablenum{ 1.0M} & \tablenum{69.82} &  $ 70.04 \pm 0.10$ & \tablenum{66.47} &  $ 66.58 \pm 0.11$ \\
\bottomrule
\end{tabular} \\
\end{table}

\begin{table}[!htb]
\caption{WikiText-2 results}
\label{table:WT2}
\begin{tabular}{c c c c c c c}
\toprule
\multirow{ 2}{*}{Model} & \multirow{ 2}{1cm}{\centering Weight \\[0.1ex] \centering Sparsity} & \multirow{2}{*}{MAC } & \multicolumn{2}{c}{Validation Perplexity} & \multicolumn{2}{c}{Test Perplexity}\\ \cmidrule(l){4-7}
  &  &  & Min &  Mean $\pm$ Std Dev &  Min &  Mean $\pm$ Std Dev \\ 
  \midrule
 \multirow{ 9}{*}{EGRU} &  
 $0\%$ &   \tablenum{7.4M } &    \tablenum{69.74} & $70.26 \pm 0.34$ &    \tablenum{66.64} & $67.21 \pm 0.28$ \\
 & $20\%$ &   \tablenum{6.0M } &    \tablenum{69.77} & $69.88 \pm 0.06$ &    \tablenum{66.71} & $66.80 \pm 0.04$ \\
 & $40\%$ &   \tablenum{4.7M } &    \tablenum{70.28} & $70.37 \pm 0.05$ &    \tablenum{67.03} & $67.11 \pm 0.06$ \\
 & $60\%$ &   \tablenum{3.3M } &    \tablenum{70.47} & $70.57 \pm 0.05$ &    \tablenum{67.12} & $67.22 \pm 0.15$ \\
 & $70\%$ &   \tablenum{2.6M } &    \tablenum{71.17} & $71.25 \pm 0.04$ &    \tablenum{67.64} & $67.76 \pm 0.07$ \\
 & $80\%$ &   \tablenum{2.0M } &    \tablenum{73.00} & $73.08 \pm 0.05$ &    \tablenum{69.40} & $69.72 \pm 0.51$ \\
 & $85\%$ &   \tablenum{1.6M } &    \tablenum{74.42} & $74.53 \pm 0.07$ &    \tablenum{70.70} & $70.85 \pm 0.09$ \\
 & $90\%$ &   \tablenum{1.1M } &    \tablenum{77.18} & $77.37 \pm 0.09$ &    \tablenum{73.33} & $73.44 \pm 0.05$ \\
 & $95\%$ &   \tablenum{0.6M } &    \tablenum{83.89} & $84.11 \pm 0.11$ &    \tablenum{79.28} & $79.44 \pm 0.12$ \\
 \midrule
 \multirow{ 9}{*}{LSTM} &  $0\%$ &   \tablenum[table-format = 2.1]{20.2M}&    68.68 & $68.85 \pm 0.12$ &    \tablenum{65.66} & $65.94 \pm 0.11$ \\
 & $20\%$ &   \tablenum[table-format = 2.1]{16.2M} &    \tablenum{68.73} & $68.82 \pm 0.06$ &    \tablenum{65.93} & $66.06 \pm 0.07$ \\
 & $40\%$ &   \tablenum[table-format = 2.1]{12.1M} &    \tablenum{68.73} & $68.88 \pm 0.10$ &    \tablenum{66.11} & $66.21 \pm 0.07$ \\
 & $60\%$ &   \tablenum{ 8.1M} &    \tablenum{69.00} & $69.22 \pm 0.11$ &    \tablenum{66.41} & $66.55 \pm 0.11$ \\
 & $70\%$ &   \tablenum{ 6.1M} &    \tablenum{69.52} & $69.78 \pm 0.14$ &    \tablenum{66.32} & $66.88 \pm 0.26$ \\
 & $80\%$ &   \tablenum{ 4.1M} &    \tablenum{70.61} & $70.86 \pm 0.11$ &    \tablenum{67.96} & $68.07 \pm 0.28$ \\
 & $85\%$ &   \tablenum{ 3.1M} &    \tablenum{71.94} & $72.11 \pm 0.08$ &    \tablenum{68.89} & $69.01 \pm 0.07$ \\
 & $90\%$ &   \tablenum{ 2.0M} &    \tablenum{74.56} & $74.69 \pm 0.08$ &    \tablenum{71.20} & $71.34 \pm 0.09$ \\
 & $95\%$ &   \tablenum{ 1.0M} &    \tablenum{86.02} & $86.06 \pm 0.03$ &    \tablenum{81.91} & $81.97 \pm 0.02$ \\
\bottomrule
\end{tabular}\\
\end{table}

\clearpage

\section{Pruning Methodology}
\label{sup:Experiments}

To achieve the best results after pruning we tested out multiple pruning techniques. We tried out parallel training and sparsification methodologies such as the lottery ticket hypothesis mentioned in \cite{Frankle2018} and the sparsify during training method mentioned in \cite{Hoefler2021}. However we achieved much better results by first training to convergence and then pruning. \citet{Hoefler2021} suggests pruning to the target sparsity in one go however we experimented with an iterative pruning method as described in sec.~\ref{sec:pruning}

\begin{figure}[!htb]
    \centering
    \includegraphics[width=0.75\textwidth]{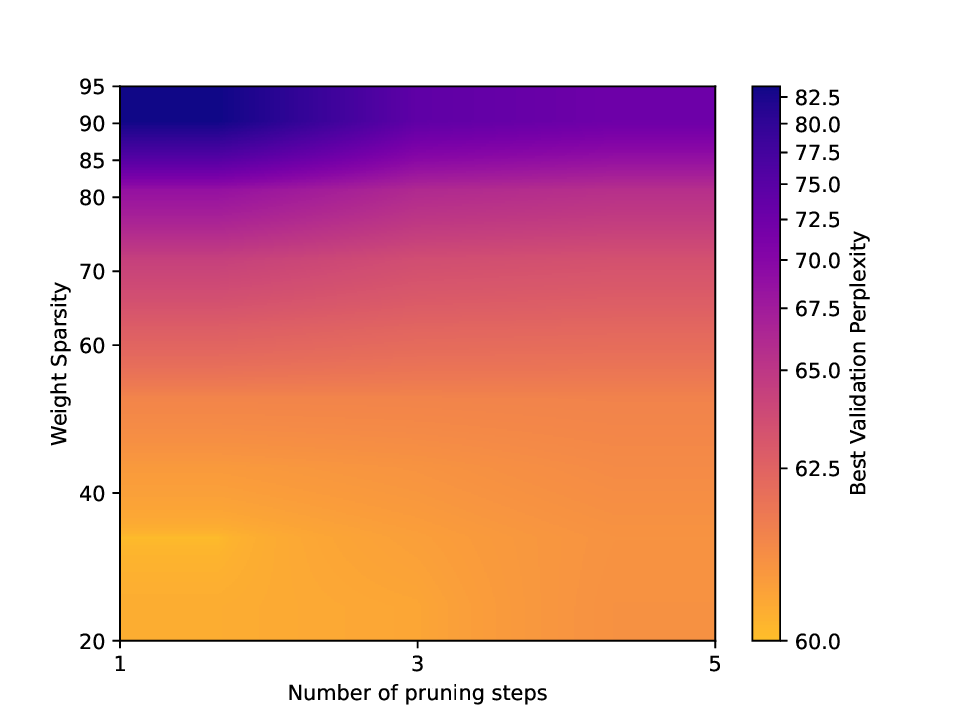}
    \caption{
     Influence of number of pruning steps on the results for different target weight sparsities. Results are displayed for EGRU experiments on Penn Treebank dataset. Other experimental setups follow similar trends
    }
    \label{fig:Seed}
\end{figure}

For the iterative pruning method, we vary the learning rate and number of pruning steps for each level of target sparsity we tried out . 
The final results are then obtained by repeating the best methodology for each target on 15 different seeds.
For low sparsity, it is beneficial to carry out the pruning in one go whereas for higher sparsity we obtain better results on carrying out multiple steps of pruning and subsequent fine-tuning.
The trends are visualized in fig.~\ref{fig:Seed}.

Similarly for learning rate, lower sparsity targets achieved better performance with a lower learning rate ($0.1\times$ baseline training learning rate) while higher sparsity models required larger learning rates (equal to the baseline training learning rate). 

\section{Network Activity}
\begin{figure}
    \centering
    \includegraphics[width=\textwidth]{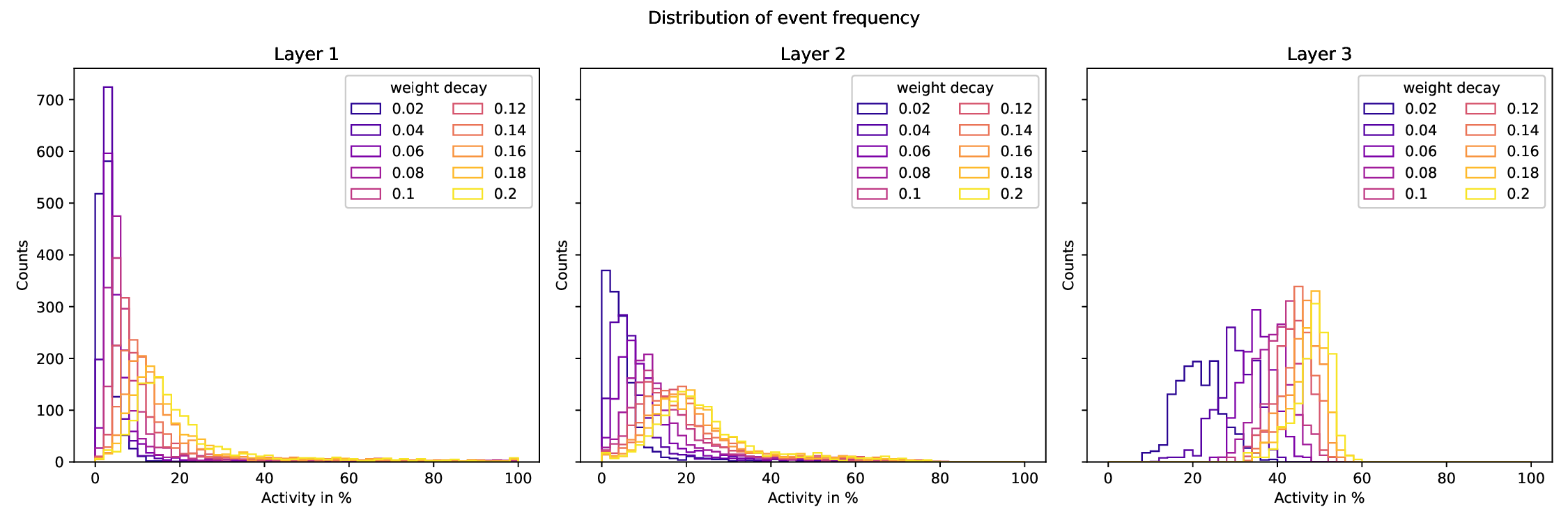}
    \caption{Histogram of activity of the EGRU neurons in each layer.
    For example, an activity of \SI{20}{\percent} denotes that a neuron's output is non-zero \SI{20}{\percent} of the time, hence saving operations \SI{80}{\percent} of the time.}
    \label{fig:frequency}
\end{figure}

Our investigation of activity sparsity in sec.~\ref{sec:activity-sparsity} showed how the distribution of weights affects the activity of EGRU networks when weight decay is applied.
We provide more detail on the activity spectrum of EGRU here.
Fig.~\ref{fig:frequency} shows the distribution of activity for the layers individually.
The lower weight decay, the more the activities move towards zero.
An anomaly is the output layer, which has significantly higher activity than the bottom layers.
This observation could be driven by the learning objective of language modeling. 
At each point in time, the model has to output a candidate word embedding vector.
This candidate vector is compared to all the (learned) word embedding vectors from the dictionary via dot-product.
Since, we don't apply a decoder layer on top of the EGRU following \citet{Merity2018}, the candidate word embeddings have positive and zero entries only.
Yet, the word embeddings are learned and perhaps take any value, especially not necessarily sparse values.
The learning objective of matching the correct next word embedding with the output of the final EGRU layer might force a high activity in the final layer.
Future work can consider more beneficial decoding strategies to keep activity low in the final layer as well.

\begin{figure}
    \centering
    \includegraphics[width=\textwidth]{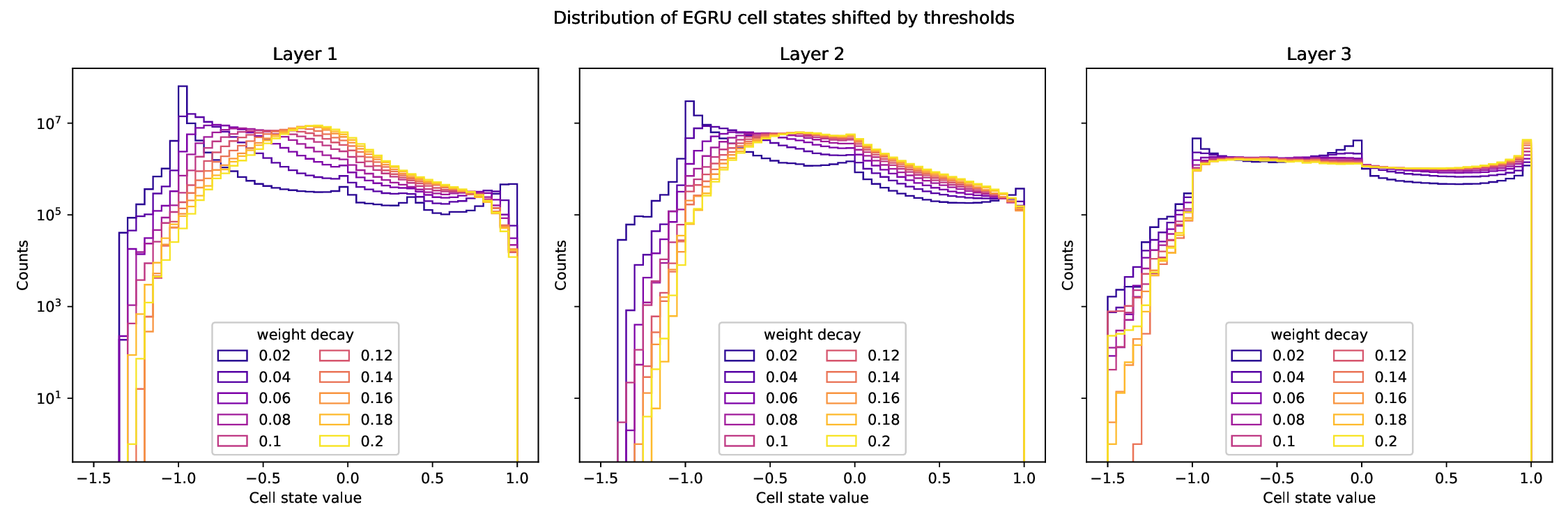}
    \caption{Histogram of cell state values shifted by the thresholds $\bfc - \bfth$ in each layer.}
    \label{fig:cell-states}
\end{figure}

Fig.~\ref{fig:cell-states} visualizes the distribution of the normalized cell states $\bfc - \bfth$ of EGRU on the Penn Treebank validation set.
We observe that higher weight decay motivates the cell states to operate closer to their thresholds.
At the same time, cells with smaller weight decay operate a significant amount of time far away from the threshold in the negative regime.
This makes the output signals $\bfy$ less sensitive to inputs, which increases the regularity of EGRU networks compared to GRU.

\section{Limitations}
Our work is based on unstructured weight sparsity and irregular activation sparsity.
Both sparsities are difficult to accelerate on contemporary hardware.
Unstructured weight sparsity is not aligned with the regular memory access instructions of GPU programming.
Yet, the sparsity pattern is known at compile time for inference applications.
This simplifies the design of specialized accelerators.
Our activation sparsity is irregular in the sense that it cannot be predicted ahead of time.
Hence, the instructions depend on context, and the compiler can hardly optimize the system e.g. by fetch weights from memory in advance.
Efficiently simulating our method requires an event-based programming paradigm, which only few accelerators such as SpiNNaker2 \citep{Mayr2019} or Loihi \citep{Loihi2018} support.    

We find that EGRU requires larger word embeddings than LSTM for the language modeling task.
This might be due to the sparsity of the feature map, which does not align with the dense embedding vectors.
The larger word embeddings introduce additional MAC operations, which limits the effective reduction of MAC operations through activity sparsity to a factor of 3 (see tab.~\ref{table:Results}).

\end{document}